\documentclass[letterpaper]{article} 
\usepackage{aaai2026}  
\usepackage{times}  
\usepackage{helvet}  
\usepackage{courier}  
\usepackage[hyphens]{url}  
\usepackage{graphicx} 
\usepackage{natbib}  
\usepackage{caption} 
\usepackage{algorithm}
\usepackage{algorithmic}
\usepackage{booktabs}
\usepackage{multirow}
\usepackage{amsmath}
\usepackage{newfloat}
\usepackage{listings}
\DeclareCaptionStyle{ruled}{labelfont=normalfont,labelsep=colon,strut=off} 
\floatstyle{ruled}
\newfloat{listing}{tb}{lst}{}
\floatname{listing}{Listing}
\title{Do LLMs Feel? Teaching Emotion Recognition with Prompts, Retrieval, and Curriculum Learning}
\author{
    Xinran Li,
    Yu Liu,
    Jiaqi Qiao,
    Xiujuan Xu\thanks{Corresponding author.}
}
\affiliations{
    School of Software Technology, Dalian University of Technology\\
    Dalian, China\\
    963707605@mail.dlut.edu.cn, yuliu@dlut.edu.cn, qiaobright@mail.dlut.edu.cn, xjxu@dlut.edu.cn
}

\begin{document}

\maketitle

\begin{abstract}
Emotion Recognition in Conversation (ERC) is a crucial task for understanding human emotions and enabling natural human-computer interaction. Although Large Language Models (LLMs) have recently shown great potential in this field, their ability to capture the intrinsic connections between explicit and implicit emotions remains limited. We propose a novel ERC training framework, \textbf{PRC-Emo}, which integrates \textbf{P}rompt engineering, demonstration \textbf{R}etrieval, and \textbf{C}urriculum learning, with the goal of exploring whether LLMs can effectively perceive emotions in conversational contexts. Specifically, we design emotion-sensitive prompt templates based on both explicit and implicit emotional cues to better guide the model in understanding the speaker’s psychological states. We construct the first dedicated demonstration retrieval repository for ERC, which includes training samples from widely used datasets, as well as high-quality dialogue examples generated by LLMs and manually verified. Moreover, we introduce a curriculum learning strategy into the LoRA fine-tuning process, incorporating weighted emotional shifts between same-speaker and different-speaker utterances to assign difficulty levels to dialogue samples, which are then organized in an easy-to-hard training sequence. Experimental results on two benchmark datasets—IEMOCAP and MELD—show that our method achieves new state-of-the-art (SOTA) performance, demonstrating the effectiveness and generalizability of our approach in improving LLM-based emotional understanding.
\end{abstract}

\begin{links}
    \link{Code}{https://github.com/LiXinran6/PRC-Emo}
\end{links}

\section{Introduction}

“Emotions play a critical role in human intelligence. For truly intelligent machines, emotional intelligence is not optional,” based on the ideas of Rosalind W. Picard, the pioneer of affective computing. As conversational agents \cite{Lee2020StudyOE} and large language models (LLMs) become increasingly integrated into our daily lives, it is essential that these systems not only understand syntax and semantics but also perceive and interpret human emotions. Emotion Recognition in Conversation (ERC) emerges as a crucial task toward building emotionally-aware AI, enabling natural and empathetic human-computer interaction \cite{hu2024recent}.

Large Language Models, with their generative architecture, have achieved significant performance improvements across various natural language processing tasks \cite{Laskar2024ASS}. Currently, the field of ERC has also entered the era of LLMs, with increasing research efforts attempting to leverage LLMs to model conversational context and predict speakers’ emotional states. By harnessing the powerful prior knowledge of LLMs, researchers incorporate contextual utterance information, speaker personality, and other background knowledge to design high-quality prompts, combined with parameter-efficient fine-tuning methods such as LoRA \cite{LoRA} for adaptive training. Although LLMs show great promise in ERC tasks, they still face the following three challenges:

\begin{figure}[t]
\centering
\includegraphics[width=0.45\textwidth]{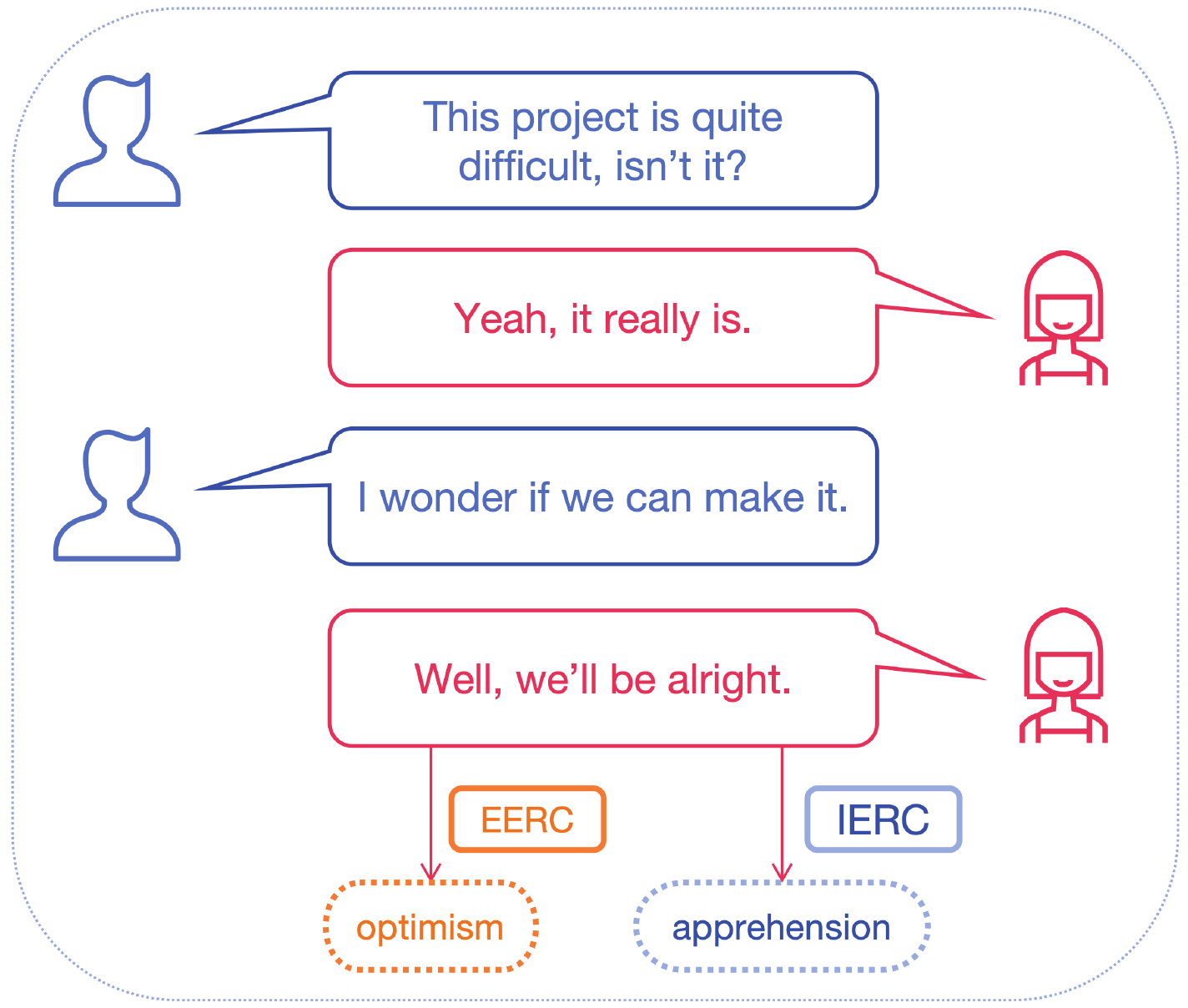} 
\caption{An example of Explicit Emotion Recognition in Conversation (EERC) and Implicit Emotion Recognition in Conversation (IERC). In this conversation, although the girl feels apprehension due to the boy’s anxious utterance, “I wonder if we can make it,” she expresses optimism in her own response to encourage both of them.}
\label{EERC}
\end{figure}

(1) Most existing studies have not adequately considered both explicit and implicit emotions when designing prompts. Emotion recognition tasks can be divided into explicit emotion recognition and implicit emotion recognition \cite{implicit}, as illustrated in Figure \ref{EERC}. Explicit emotion recognition focuses on predicting emotions conveyed by the speaker through direct expressions, strong intonations, or obvious facial cues, while implicit emotion recognition aims to capture emotions genuinely felt by the speaker but not necessarily expressed through language. Ignoring the balance between these two aspects limits the model’s ability to fully comprehend complex emotional states, thereby constraining the accuracy of emotion recognition.

(2) Current research often employs demonstration retrieval in prompt design, extracting the most similar sentences and their corresponding labels from a demonstration repository to assist the model in understanding and predicting emotions \cite{InstructERC}. However, existing methods mostly construct the demonstration repository using only the training sets of commonly used datasets, which limits the diversity and coverage of demonstration samples. Due to the relatively homogeneous content of the repository, the model's reasoning is often confined to the context of specific datasets, lacking sufficient generalization ability. This restricts the model’s adaptability to more complex and diverse conversational scenarios, thereby limiting its practical effectiveness.

(3) At present, most studies mainly focus on optimizing prompt design and fine-tuning strategies, with relatively little attention paid to improving the overall training process \cite{HybridCL}. These methods often involve complex prompt designs and multi-stage training strategies, resulting in high computational resource consumption but limited performance gains. Moreover, ERC datasets generally suffer from severe class imbalance, making it difficult for models to adequately learn the features of low-frequency categories, which in turn affects overall performance and generalization ability. Therefore, optimizing training strategies to enhance model performance on imbalanced datasets remains an urgent and important challenge.

To address the above challenges, we propose PRC-Emo: a novel Prompt-Retrieval-Curriculum framework for ERC. Leveraging LLMs, our method takes both the current utterance and its dialogue history as input to generate two types of emotion interpretations: explicit (directly expressed emotions) and implicit (underlying, unspoken emotions). These complementary signals are integrated into structured prompts, enabling the model to better capture the nuanced emotional states of speakers. We further introduce the first ERC-specific demonstration retrieval repository, which combines high-quality samples from multiple datasets (IEMOCAP, MELD, EmoryNLP) and over 10,000 additional utterances generated by LLMs and refined by human annotators. These samples span six real-world domains—healthcare, workplace, education, family, social, and entertainment—offering greater diversity and contextual richness for few-shot prompting. We improve the Curriculum Learning strategy by defining a dialogue difficulty metric that combines weighted emotional shifts both within the same speaker and between different speakers. Based on this metric, we divide the training data into multiple levels (i.e., "training buckets") from easy to hard, enabling the model to learn progressively. The main contributions are:

\begin{itemize}
\item We propose a method that generates explicit and implicit emotion interpretations as external knowledge in prompts, enabling a better capture of the speaker's true emotional state.
\item We construct the first demonstration retrieval repository for ERC that combines multi-source data and LLM-generated, human-refined samples, significantly enhancing the model’s generalization ability.
\item We enhance a curriculum learning strategy based on dialogue difficulty, using dynamic training buckets for progressive learning and better robustness.
\item Experiments on multiple benchmarks show our method achieves consistent SOTA results. The code and data have already been released on GitHub.
\end{itemize}

\begin{figure*}[t]
\centering
\includegraphics[width=1\textwidth]{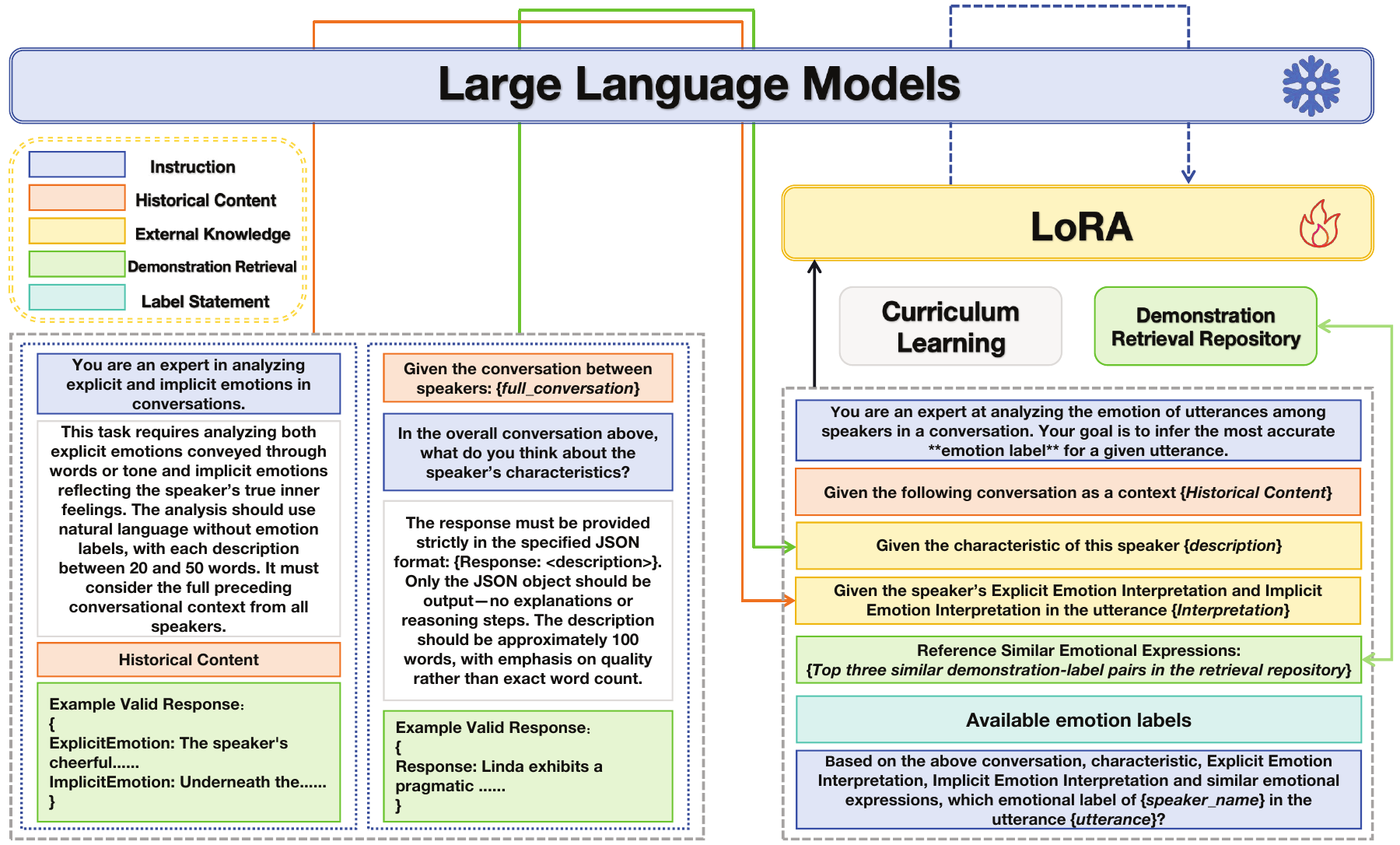} 
\caption{PRC-Emo’s architecture has two main stages: extracting external supplementary knowledge and predicting emotion labels, with curriculum learning applied during training. The two prompts at the bottom-left extract explicit and implicit emotion interpretations and speaker characteristics as external knowledge. This information is passed to the bottom-right prompt, which performs the final emotion recognition by retrieving similar pairs from a retrieval repository to aid the process.}
\label{framework}
\end{figure*}

\section{Related Work}

With the rise of LLMs, the field of ERC can now be broadly divided into two categories: traditional methods and LLM-based methods. This section provides an overview of the methods previously used, as well as the application of curriculum learning in ERC.

\subsection{Traditional Methods}
Traditional research in the field of ERC has mainly focused on three directions: Recurrent Neural Networks (RNNs), Graph Neural Networks (GNNs), and Pre-trained Language Models (PLMs).

In RNN-based studies, models typically leverage RNNs to perform sequential modeling of utterances or entire conversations. DialogueRNN \citep{DialogueRNN} utilizes an RNN structure to dynamically track the individual state of each speaker throughout the conversation. DialogueCRN \citep{DialogueCRN} builds upon this by incorporating cognitive factors to enhance the understanding of both contextual and speaker-level information. In GNN-based studies, models construct graph structures to capture dependencies between utterances and speakers. DialogueGCN \citep{DialogueGCN} treats utterances as nodes in a graph and builds edges based on contextual information to represent semantic relationships. DAG-ERC \citep{DAG-ERC} further considers speaker identity and utterance position when constructing a directed acyclic graph to model dialogue structures more precisely. In the area of pre-trained language models, models such as BERT \cite{BERT} and RoBERTa \cite{RoBERTa} have been widely adopted due to their powerful language representation capabilities. BERT-ERC \cite{BERT-ERC} enhances model performance by introducing guided texts, fine-grained emotion classification modules, and a two-stage training strategy.

\subsection{LLM-based Methods}
Traditional ERC methods predominantly adopt discriminative architectures. With the rise of LLMs, InstructERC \citep{InstructERC} is the first to introduce a generative architecture for tackling ERC tasks, opening up a novel research direction. This approach incorporates an emotion-alignment auxiliary task and a retrieval-augmented prompting module to enhance the model’s understanding of emotions. Moreover, BiosERC \cite{BiosERC} enriches prompts with background information, such as speaker characteristics, and applies LoRA for the efficient fine-tuning of large models. These LLM-based approaches have significantly improved performance across various ERC benchmarks, demonstrating powerful generalization and adaptability, and paving the way for future advancements in the field.

\subsection{Curriculum Learning}
Curriculum Learning (CL) \cite{CL}, a training strategy that simulates human learning, has gained widespread attention in ERC in recent years. \citet{HybridCL} proposed a Hybrid Curriculum Learning framework that combines difficulty designs at both the conversation and utterance levels, guiding the model to gradually learn complex emotions based on the frequency of emotion shifts and emotional similarity. \citet{curriculum-2024} introduced the MultiDAG+CL method, which integrates textual, acoustic, and visual features and leverages curriculum learning to address challenges related to emotional variations and data imbalance. \citet{LSDGNN} introduce the concept of weighted emotion shifts, with a difficulty design that focuses on modeling transitions between similar emotions. However, these methods only consider emotion transitions within the same speaker and do not take into account transitions between different speakers.

\section{Methodology}
This section presents the PRC-Emo architecture. We first define the problem, then describe the extraction of explicit and implicit emotion interpretations as external knowledge. Next, we detail the prompt retrieval template module and its ERC-specific demonstration retrieval repository. Finally, we explain the curriculum learning strategy used in training. The overall architecture of the PRC-Emo is illustrated in Figure \ref{framework}.

\subsection{Problem Definition}
In Emotion Recognition in Conversation, a conversation is represented as a sequence of utterances \( D = \{u_1, u_2, u_3, ..., u_N\} \), where \( N \) denotes the total number of utterances. Each utterance \( u_{i,s_j} \) is spoken by a specific speaker \( s_j \), indicating that the \( i^{th} \) utterance is from speaker \( s_j \). The objective of ERC is to assign an emotion label \( y_k \in Y \), such as joy or sadness, to each utterance \( u_i \), where \( Y \) is the set of possible emotion categories.

\subsection{Extraction of External Supplementary Knowledge}
Given the extensive knowledge and powerful language understanding abilities of LLMs, we adopt a Prompt Engineering approach by designing query templates to extract supplementary information valuable for ERC tasks. Inspired by \citet{implicit}, we observe that the current ERC field generally lacks clear definitions and systematic interpretations of explicit emotion and implicit emotion. Traditional models typically infer the emotion of an utterance based solely on the textual content of the current and historical utterances, without truly understanding the speaker’s internal emotional state.

To address this, we design a set of query templates specifically for generating explicit and implicit emotion interpretations. These interpretations serve as high-quality auxiliary signals and are injected into the model input, guiding the model to attend to both the speaker’s expressed emotions and their underlying emotional state. This enhances the model’s ability to capture multi-layered emotional expressions. In addition, inspired by \citet{BiosERC}, we also extract speaker characteristic as supplementary knowledge to further support emotion recognition. These external supplementary knowledge sources are incorporated into the Retrieval Template Module for the final emotion recognition. For the interpretations of explicit emotion and implicit emotion, we input the historical utterances. For the speaker characteristic, we use the entire dialogue. The specific prompt design is shown in the bottom-left corner of Figure \ref{framework}.

\subsection{Retrieval Template Module}
To better leverage LLMs, we construct a carefully designed Retrieval Template Module and fine-tune the model using LoRA technology. In this section, we introduce the components of the Retrieval Template Module, as well as the first demonstration retrieval repository specifically built for the ERC task. The specific prompt design is shown in the bottom-right corner of Figure \ref{framework}.

\subsubsection{Components of Retrieval Template Module}

Each input consists of an Instruction, Historical Content, External Knowledge, Demonstration Retrieval, and a Label Statement.

\begin{itemize}
    \item \textbf{Instruction}. Defines the model's role and core task objectives.
    \item \textbf{Historical Content}. In order to determine the emotion of a given utterance, it is necessary to provide the model with historical context. We adopt a history window $w$, which includes the past and current $w$ utterances along with the corresponding speaker names. 
    \item \textbf{External Knowledge}. It consists of two parts: speaker characteristic descriptions and expert-level interpretations of explicit/implicit emotions. The speaker characteristics and expert emotion interpretations vary across different conversations.
    \item \textbf{Demonstration Retrieval}. Numerous studies have demonstrated the importance of demonstration retrieval \cite{Retrieved_Demonstrations}. To extract utterances most similar to the current one, we select three ERC examples from our self-constructed demonstration retrieval repository $D_{domain}$. We utilize $SBERT$ \citep{SBERT} to retrieve relevant demonstrations from the repository. By calculating the cosine similarity and comparing with all vectors in $D_{domain}$, the top three demonstration-label pairs with the highest scores are selected.
    \item \textbf{Label Statement}. We constrain the model output to a limited set of labels based on the specific dataset used.
\end{itemize}

\subsubsection{Demonstration Retrieval Repository}
To address the issue of imbalanced emotion label distribution in emotion recognition tasks, we construct an emotion dialogue augmentation dataset based on OpenAI's GPT-4o, covering a wide range of real-life scenarios. The dataset spans common contexts such as healthcare, workplace, education, family, social interactions, and entertainment, and includes five emotion categories: happiness, neutral, fear, sadness, and anger.

We adopt a two-stage prompt strategy for data generation. First, the model produces 30 subtopics under a given scenario to increase diversity. Then, it generates coherent two-person dialogues for these subtopics, with each utterance assigned an emotion label. This process enables targeted creation of underrepresented emotions while preserving contextual consistency, resulting in a more balanced and diverse corpus for downstream ERC models.

For annotation, we adopt a “label masking + human verification” strategy. GPT-4o first generates text samples with emotion labels, after which the labels are removed. Then, two researchers independently annotate each utterance with an emotion category. A sample is officially included in the augmented dataset only if both annotators’ judgments exactly match the original label. After one round of filtering, we identify emotion categories still lacking sufficient samples and generate additional data accordingly. This process of generation and filtering repeats three times to finally obtain a high-quality emotion dataset with the desired distribution. Appendix A provides the detailed composition of our self-constructed dataset along with example sentences.

Based on our self-constructed dataset and the training sets of three widely-used ERC benchmarks—IEMOCAP, MELD, and EmoryNLP—we build the first demonstration retrieval repository in the ERC domain. This repository contains a total of 36,712 utterances, including 14,009 from our dataset, 5,163 from IEMOCAP, 9,989 from MELD, and 7,551 from EmoryNLP. Each utterance in the repository includes the following information: the original text, emotion label, source dataset, dialogue ID, sentence position within the dialogue, and its vector representation. By default, we use $SBERT$ to encode the utterances into high-quality semantic embeddings. This demonstration retrieval repository provides a unified, structured, and high-quality data foundation for future research on emotion-aware retrieval and prompt engineering.

\subsection{Curriculum Learning}
Curriculum learning \citep{CL} trains models from easy to hard. This section presents our difficulty measure and training scheduler.

\subsubsection{Difficulty Measure Function}
Previous studies \citep{curriculum-2024, LSDGNN} use emotional shifts as difficulty metrics but overlook transitions between different speakers. We address this by introducing a difficulty function based on weighted emotional shift frequency that accounts for emotional similarity, enabling more accurate conversation complexity estimation.

The weighted emotional shift is defined as the emotional change between two consecutive utterances. Based on the two-dimensional arousal-valence emotion wheel, each emotion label corresponds to a point on the unit circle, covering all emotion categories in the ERC dataset. The similarity between different emotions is calculated by Equation \eqref{eq:similar}:
\begin{eqnarray}
s_{ij} =
\begin{cases}
\max(\cos(\theta_{ij}), 0) & \text{if } \mathbf{v}_i \cdot \mathbf{v}_j > 0 \\
0 & \text{if } \mathbf{v}_i \cdot \mathbf{v}_j < 0 \\
\frac{1}{N} & \text{if } \mathbf{v}_i \cdot \mathbf{v}_j = 0
\end{cases}
\label{eq:similar}
\end{eqnarray}
Here, $s_{ij}$ denotes the similarity between emotion labels $i$ and $j$, $\mathbf{v}_i$ represents the valence vector of label $i$, $\theta_{ij}$ is the angle between labels $i$ and $j$, and $N$ is the total number of emotion labels in the dataset. The closer two emotions are, the higher their similarity score. The weighted emotional shift (WES) is then defined as shown in Equation \eqref{eq:simi}:
\begin{eqnarray}
\label{eq:simi}
N^{WES} = k \times s_{ij} + b
\end{eqnarray}
We apply a linear transformation to $s_{ij}$, where $k$ is a weighting factor and $b$ is a bias term. Thus, the difficulty of a conversation $c_i$ is defined as shown in Equation \eqref{eq:suanfa2}, \eqref{eq:suanfa3} and \eqref{eq:suanfa4}:

\begin{eqnarray}
DIF(c_i) = \frac{WES_{same}(c_i) + WES_{diff}(c_i)
 + N_{sp}(c_i)}{N_{u}(c_i) + N_{sp}(c_i)} 
\label{eq:suanfa2}
\end{eqnarray}
\begin{eqnarray}
WES_{same}(c_i) = \sum_{j=1}^{N_{shift}^{same}(c_i)}N_j^{WES} 
\label{eq:suanfa3}
\end{eqnarray}

\begin{eqnarray}
WES_{diff}(c_i) = \sum_{j=1}^{N_{shift}^{diff}(c_i)}N_j^{WES}
\label{eq:suanfa4}
\end{eqnarray}
where $N_{shift}^{same}(c_i)$ and $N_{shift}^{diff}(c_i)$ represents the total number of emotional shifts between utterances from the same speaker and between consecutive utterances from different speakers in conversation $c_i$. $N_{u}(c_i)$ is the total number of utterances. $N_{sp}(c_i)$ denotes the number of speakers appearing in conversation $c_i$, which serves as a smoothing factor. $N_j^{WES}$ is the Weighted Emotional Shift at the $j$-th emotional shift. The proposed algorithm is presented in Algorithm \ref{alg:cl_training}.

\begin{algorithm}[H]
\caption{Curriculum Learning Training with Difficulty Measure Function}
\label{alg:cl_training}
\begin{algorithmic}[1]
\REQUIRE 
\STATE \textbf{Input:} Training dataset $D$, Model $M$, Number of buckets $n$, Training epochs $t$
\STATE \textbf{Parameters:} Linear transformation coefficients $(k, b)$ for WES calculation
\ENSURE Trained model $M^*$

\STATE \textbf{// Phase 1: Calculate difficulty for each conversation}
\FOR{each conversation $c_i \in D$}
    \STATE Initialize: $WES_{same} \leftarrow 0$, $WES_{diff} \leftarrow 0$, $N_{sp} \leftarrow 0$, $N_u \leftarrow 0$
    \STATE $S \leftarrow \emptyset$ \COMMENT{Speaker emotion sequences}
    
    \STATE \textbf{// Build speaker emotion sequences}
    \FOR{each utterance $u_j$ in $c_i$}
        \STATE $speaker\_id \leftarrow \text{get\_speaker}(u_j)$
        \STATE $emotion \leftarrow \text{get\_emotion}(u_j)$
        \STATE $S[speaker\_id] \leftarrow S[speaker\_id] \cup \{emotion\}$
        \STATE $N_u \leftarrow N_u + 1$
    \ENDFOR
    \STATE $N_{sp} \leftarrow |S|$ \COMMENT{Number of unique speakers}
    
    \STATE \textbf{// Calculate same speaker emotional shifts}
    \FOR{each speaker $p \in S$}
        \FOR{$j = 1$ to $|S[p]| - 1$}
            \IF{$S[p][j] \neq S[p][j+1]$}
                \STATE $s \leftarrow \text{similarity}(S[p][j], S[p][j+1])$
                \STATE $WES_{same} \leftarrow WES_{same} + (k \times s + b)$
            \ENDIF
        \ENDFOR
    \ENDFOR
    
    \STATE \textbf{// Calculate different speaker emotional shifts}
    \FOR{$j = 1$ to $N_u - 1$}
        \IF{$\text{speaker}(u_j) \neq \text{speaker}(u_{j+1})$}
            \STATE $s \leftarrow \text{similarity}(\text{emotion}(u_j), \text{emotion}(u_{j+1}))$
            \STATE $WES_{diff} \leftarrow WES_{diff} + (k \times s + b)$
        \ENDIF
    \ENDFOR
    
    \STATE \textbf{// Compute conversation difficulty}
    \STATE $DIF(c_i) \leftarrow \frac{WES_{same} + WES_{diff} + N_{sp}}{N_u + N_{sp}}$
\ENDFOR

\STATE \textbf{// Phase 2: Sort and partition dataset}
\STATE $D_{sorted} \leftarrow \text{sort}(D, \text{by } DIF \text{ ascending})$
\STATE Partition $D_{sorted}$ into $n$ buckets: $\{D_1, D_2, \ldots, D_n\}$
\STATE where $\max(DIF(D_i)) \leq \min(DIF(D_{i+1}))$ for $i = 1, \ldots, n-1$

\STATE \textbf{// Phase 3: Curriculum training}
\STATE $D_{train} \leftarrow \emptyset$
\FOR{$epoch = 1$ to $t$}
    \IF{$epoch \leq n$}
        \STATE $D_{train} \leftarrow D_{train} \cup D_{epoch}$ \COMMENT{Add next difficulty bucket}
    \ENDIF
    \STATE $M \leftarrow \text{train}(M, D_{train})$
\ENDFOR

\RETURN $M^*$
\end{algorithmic}
\end{algorithm}

\subsubsection{Training Scheduler}
The training scheduler aims to divide the entire dataset $D$ into several subsets with similar difficulty levels (i.e., “training buckets”), denoted as $\{D_1, D_2, ..., D_n\}$. The model training starts with the easiest subset. After completing several epochs, the next more difficult subset is gradually introduced. Once all subsets have been involved in training, additional epochs are conducted on the entire dataset to further improve model performance.

\section{Experimental Settings}
This section introduces the datasets, baselines, and implementation details.

\subsection{Datasets}
We use two ERC datasets: IEMOCAP \citep{IEMOCAP}, which consists of dyadic conversations; and MELD \citep{MELD}, a multiparty conversation dataset derived from \textit{Friends}. The composition of datasets is shown in Table \ref{dataset}.

\begin{table}[t]
\centering
\begin{tabular}{cccc}
\toprule
Dataset & Partition &  Utterance & Dialogues \\
\midrule
\multirow{2}{*}{IEMOCAP} & train + val & 5810 & 120 \\
& test & 1623 & 31 \\
\multirow{2}{*}{MELD} & train + val & 11098 & 1152 \\
& test & 2610 & 280 \\
\bottomrule
\end{tabular}
\caption{Statistics of the two datasets.}
\label{dataset}
\end{table}

\subsection{Baselines}
We compare our proposed method with two categories of mainstream baselines: (1) Conventional models, including DialogueRNN \cite{DialogueRNN}, ICON \cite{ICON}, DialogueGCN \cite{DialogueGCN}, COSMIC \cite{COSMIC}, MMGCN \cite{MMGCN}, DAG-ERC \cite{DAG-ERC}, LR-GCN \cite{LR-GCN}, MultiDAG+CL \cite{curriculum-2024}, CBERL \cite{CBERL}, DER-GCN \cite{DER-GCN} and LSDGNN+ICL \cite{LSDGNN}; and (2) Large language model (LLM)-based methods, including InstructERC \cite{InstructERC} and BiosERC \cite{BiosERC}. Following the convention in the ERC field, the evaluation metric for most comparative experiments in this paper is the weighted F1 score by default.

\subsection{Implementation Details}
All experiments are conducted on a single NVIDIA 4090D GPU. For the IEMOCAP dataset, Qwen2.5-7B-Instruct is used as the base model; for the MELD dataset, Qwen3-8B serves as the base model. The external large language model is consistently Qwen3-14B, which is employed to generate explicit and implicit emotion interpretations as well as speaker characteristics. Additional details on training hyperparameters and settings are provided in Appendix B. All experiments in this paper—including comparative experiments—are repeated using five different random seeds, and the average results are reported. 


\section{Results and Analysis}
This section presents the comparison between our model and state-of-the-art methods, the results of ablation studies, and a series of comparative experiments.
\subsection{Comparison with the State of the Art}
Table \ref{results} presents the performance of our model on the IEMOCAP and MELD, with bold values indicating the best results among all models. The results of other models are taken from their original papers. Our method, PRC-Emo, achieves SOTA performance on both the IEMOCAP and MELD datasets, with weighted F1 score improvements of 0.76\% on IEMOCAP and 0.61\% on MELD. As a supplement, we also compare accuracy scores and achieve the best results.

\begin{table*}[t]
\centering
\begin{tabular}{l|cc|cc|cc}
\toprule
\multirow{2}{*}{\textbf{Model}} & \multicolumn{2}{c|}{\textbf{IEMOCAP}} & \multicolumn{2}{c|}{\textbf{MELD}} & \multicolumn{2}{c}{\textbf{Average}} \\
\cmidrule(lr){2-3} \cmidrule(lr){4-5} \cmidrule(l){6-7}
 & \textbf{Acc.} & \textbf{W-F1} & \textbf{Acc.} & \textbf{W-F1} & \textbf{Acc.} & \textbf{W-F1} \\
\midrule
DialogueRNN \citep{DialogueRNN} & 63.40 & 62.75 & 60.27 & 57.03 & 61.84 & 59.89 \\
ICON \citep{ICON} & 64.00 & 63.50 & --- & 56.30 &--- & 59.90 \\
DialogueGCN \citep{DialogueGCN} & 65.25 & 64.18 & --- & 58.10 & --- & 61.14 \\
COSMIC \cite{COSMIC} & --- & 65.28 & --- & 65.21 & --- & 65.25 \\
MMGCN \citep{MMGCN} & --- & 66.22 & 61.34 & 58.65 & --- & 62.44 \\
DAG-ERC \citep{DAG-ERC} & 67.53 & 68.03 & 63.98 & 63.63 & 65.76 & 65.83 \\
LR-GCN \citep{LR-GCN} & 68.50 & 68.30 & --- & 65.60 & --- & 66.95 \\
MultiDAG+CL \cite{curriculum-2024} & 69.11 & 69.08 & --- & 64.00 & --- & 66.54 \\
CBERL \citep{CBERL} & 69.36 & 69.27 & 67.78 & 66.89 & 68.57 & 68.08 \\
DER-GCN \citep{DER-GCN} & 69.70 & 69.40 & 66.8 & 66.10 & 68.25 & 67.75 \\
LSDGNN+ICL \citep{LSDGNN} & 70.35 & 70.24 & 64.67 & 64.07 & 67.51 & 67.16 \\
InstructERC \cite{InstructERC} & --- & 71.39 & --- & 69.15 & --- & 70.27 \\
BiosERC \cite{BiosERC} & --- & 71.19 & --- & 69.83 & --- & 70.51 \\
\midrule
\textbf{PRC-Emo (Ours)} & \textbf{71.03} & \textbf{71.95} & \textbf{71.50} & \textbf{70.44} & \textbf{71.27} & \textbf{71.20} \\
\bottomrule
\end{tabular}
\caption{Performance comparison of different ERC methods on IEMOCAP and MELD datasets.}
\label{results}
\end{table*}

\subsection{Ablation Experiments}
To investigate the importance of the \textbf{P}rompt, demonstration \textbf{R}etrieval, and \textbf{C}urriculum Learning modules in the PRC-Emo model, we conduct ablation experiments on two datasets. The results are shown in Table \ref{ablation}.
\begin{table}[t]
\centering
\begin{tabular}{lcc}
        \toprule
        Model & IEMOCAP & MELD \\
        \midrule
        \textbf{PRC-Emo} & \textbf{71.95} & \textbf{70.44} \\
        w/o C & 71.52 (↓ 0.43) & 70.07 (↓ 0.37) \\
        w/o R + C & 70.74 (↓ 1.21) & 69.62 (↓ 0.82) \\
        w/o P + R + C & 68.54 (↓ 3.41) & 68.72 (↓ 1.72) \\
        \bottomrule
\end{tabular}
\caption{Ablation experiments.}
\label{ablation}
\end{table}

From the results, it can be seen that each module plays an important role. Among them, the design of the Prompt contributes the most significant improvement, which demonstrates that the interpretations of explicit and implicit emotions have a remarkable effect. This enables the model to deeply understand the speaker’s psychological state and make better predictions. Demonstration retrieval and curriculum learning further enhance the performance. When the P, R, and C modules are removed, the model essentially becomes equivalent to directly fine-tuning LLM using LoRA.

\subsection{Comparative Experiments on Prompt Design}
To verify the effectiveness of our prompt design, we conduct comparative ablation experiments on different components of the prompt. Our prompt augmentation consists of the following components: \textbf{S}peaker  characteristics, explicit emotion and implicit emotion \textbf{I}nterpretations, and demonstration \textbf{R}etrieval. The training process adopts curriculum learning. The results are shown in Table \ref{prompt}.
\begin{table}[t]
\centering
\begin{tabular}{lcc}
        \toprule
        Model & IEMOCAP & MELD \\
        \midrule
        \textbf{PRC-Emo} & \textbf{71.95} & \textbf{70.44} \\
        w/o R & 71.49 (↓ 0.46) & 70.23 (↓ 0.21) \\
        w/o I + R & 70.27 (↓ 1.68) & 69.66 (↓ 0.78) \\
        w/o S + I + R & 69.90 (↓ 2.05) & 69.34 (↓ 1.10) \\
        \bottomrule
\end{tabular}
\caption{Prompt design comparative experiments.}
\label{prompt}
\end{table}

Table \ref{prompt} shows that the full prompt design (PRC-Emo) achieves the best performance. Gradually removing demonstration retrieval (R), explicit emotion and implicit emotion interpretations (I), and speaker characteristics (S) leads to a consistent performance drop, indicating that each component contributes positively to the model. Among them, explicit emotion and implicit emotion interpretations (I) have the most significant impact.

\subsection{Comparative Experiments on Curriculum Learning Methods for ERC}
\citet{curriculum-2024} proposed using emotional shifts (ES) as the key metric for measuring sentence difficulty. \citet{LSDGNN} proposed using weighted emotional shifts (WES) as the key metric. Both methods only consider emotional changes within the same speaker. Our curriculum learning approach introduces emotional changes across different speakers, providing a better assessment of sentence difficulty. Table \ref{cl} presents the performance comparison of the three curriculum learning methods. All comparisons are performed using the same base model and prompt, with ES and WES representing their respective curriculum learning strategies.

\begin{table}[t]
\centering
\begin{tabular}{lcc}
        \toprule
        Methods & IEMOCAP & MELD \\
        \midrule
        base + Ours & \textbf{71.95} & \textbf{70.44} \\
        base + ES  & 70.43 (↓ 1.52) &  69.87(↓ 0.57) \\
        base + WES & 71.48 (↓ 0.47) &  70.14(↓ 0.30) \\
        \bottomrule
\end{tabular}
\caption{Comparison of curriculum learning methods.}
\label{cl}
\end{table}

\subsection{Comparative Experiments of Different LLMs}
Due to resource constraints, we only compare two of the latest high-performing open-source large language models: Qwen2.5-7B and Qwen3-8B. The experimental results are shown in Table \ref{llm}. We observe an interesting phenomenon: the performance of different LLMs varies significantly across datasets. Specifically, Qwen2.5-7B significantly outperforms Qwen3-8B on the IEMOCAP dataset, while Qwen3-8B performs significantly better than Qwen2.5-7B on the MELD dataset. This may be due to Qwen3's optimization for complex real-world dialogues and possible data leakage, as MELD is based on the widely available script of the TV series \textit{Friends}.
\begin{table}[t]
\centering
\begin{tabular}{lcc}
        \toprule
        Model & IEMOCAP & MELD \\
        \midrule
        Qwen2.5-7B & \textbf{71.95} & 69.73 \\
        Qwen3-8B  & 70.86   & \textbf{70.44} \\
        \bottomrule
\end{tabular}
\caption{Comparison of different LLMs.}
\label{llm}
\end{table}

\section{Conclusion}
This paper proposes PRC-Emo, a novel training framework for Emotion Recognition in Conversation (ERC) that integrates \textbf{P}rompt engineering, demonstration \textbf{R}etrieval, and \textbf{C}urriculum learning using large language models (LLMs). PRC-Emo builds the first dedicated demonstration retrieval repository for ERC and designs emotion-sensitive prompt templates based on explicit and implicit emotional cues to better understand psychological states. The curriculum learning strategy organizes training from easy to hard based on weighted emotional shifts between same-speaker and different-speaker utterances. Experimental results on the IEMOCAP and MELD benchmark datasets show that PRC-Emo achieves new state-of-the-art performance. This work highlights the potential of combining prompt-based learning with curriculum learning strategies to advance ERC tasks. In future work, we plan to further explore stronger LLMs, more efficient prompting paradigms, and advanced curriculum designs to enhance emotional reasoning and improve the robustness and generalization of ERC systems.

\section*{Acknowledgments}
This work is funded in part by the National Natural Science Foundation of China Project (No. 62372078).

\bibliography{aaai2026}
\appendix
\section{Appendix A}
\label{appendix:dataset}
Our self-constructed dataset consists of 14,009 sentences spanning seven distinct domains: healthcare, workplace, education, family, social interactions, entertainment, and comprehensive scenarios. The dataset maintains a balanced distribution across both emotional categories and domain contexts. Among the domains, the comprehensive category contains the largest number of samples (3,314 sentences), followed by family interactions (2,015 sentences) and workplace scenarios (1,862 sentences). The healthcare domain represents the smallest category with 1,625 sentences. Across emotional categories, neutral emotions are most prevalent (3,309 sentences), while anger represents the least frequent category (2,566 sentences). This multi-domain, balanced distribution ensures comprehensive coverage of real-world conversational scenarios while providing sufficient samples for robust model training across diverse contextual settings. As shown in Table \ref{domain_emotion_distribution}, the distribution of data samples and emotion labels across each category is presented.

\begin{table*}[b]
\centering
\begin{tabular}{lcccccc}
\toprule
Domain & Happiness & Neutral & Fear & Sadness & Anger & Total \\
\midrule
Healthcare & 171 & 421 & 345 & 381 & 307 & 1,625 \\
Workplace & 155 & 550 & 398 & 459 & 300 & 1,862 \\
Education & 486 & 324 & 245 & 276 & 406 & 1,737 \\
Family & 147 & 562 & 392 & 451 & 463 & 2,015 \\
Social & 518 & 341 & 320 & 312 & 297 & 1,788 \\
Entertainment & 613 & 332 & 281 & 267 & 175 & 1,668 \\
Comprehensive & 623 & 779 & 642 & 652 & 618 & 3,314 \\
\midrule
Total & 2,713 & 3,309 &2,623 &2,798 & 2,566 &14,009 \\
\bottomrule
\end{tabular}
\caption{Distribution of emotions across different domains in the self-constructed dataset.}
\label{domain_emotion_distribution}
\end{table*}

Table \ref{dataset_examples_part1} and Table \ref{dataset_examples_part2} provide representative examples from different domains in our dataset, illustrating the diversity and contextual richness of the collected dialogues. 

\begin{table*}[htbp]
\centering
\begin{tabular}{l|p{12cm}c}
\toprule
Domain & Example Utterance & Emotion\\
\midrule
Healthcare 
&(1) I've been feeling great lately! I just had a checkup, and the doctor said everything looks good! & Happiness \\
&(2) That's fantastic! Do you think the checkup process was complicated at all? & Neutral \\
&(3) Not really! There was just a little bit of waiting, but it was totally worth it! & Happiness \\
&(4) I'm a bit worried about my own checkup results. I've heard some people get some pretty scary diagnoses. & Fear \\
&(5) I completely understand. Waiting for those results can be really stressful, but regular checkups are so important. They help catch any issues early on. & Neutral \\
&(6) Exactly! It's all about staying positive. Even if the results aren't great, knowing is better than being in the dark! & Happiness \\
&(7) You're right! I think I need to schedule my checkup too and stop worrying so much. & Neutral \\
&(8) Definitely! Facing it is way better than hiding from it! & Neutral \\
\midrule
Workplace 
&(1) Lately, I feel like it's becoming harder to find a balance between work and life, and it's really frustrating! & Anger \\
&(2) I feel the same way. I often work late, and by the time I get home, I'm completely exhausted and have no time to spend with my family. & Sadness \\
&(3) That scares me. I'm worried that if this continues, it will affect my health and my relationships! & Fear \\
&(4) Exactly, work takes up so much of my time that I'm starting to wonder if I'm wasting my life. & Sadness \\
&(5) Sometimes, I seriously think about quitting my job to pursue something more meaningful! & Anger \\
&(6) But I'm trying to figure out solutions, like setting work boundaries and scheduling regular breaks. & Neutral \\
&(7) That sounds great! I should try that too and give myself some space! & Happiness \\
&(8) Yes, finding a balance that works for us is essential for living a more fulfilling life! & Happiness \\
\midrule
Education 
&(1) I really believe that parents play a crucial role in their children's education; their support can make kids more confident. & Happiness \\
&(2) However, sometimes I feel helpless because parents' expectations can put a lot of pressure on children. & Sadness \\
&(3) Yeah, high expectations can really make kids feel scared; they worry that they won't be good enough. & Fear \\
&(4) That's why communication is so important; parents need to understand their children's feelings rather than just chase grades. & Neutral \\
&(5) I hope more parents realize that companionship and understanding are more important than grades. & Anger \\
&(6) Exactly! Education isn't just about teaching kids knowledge; it's also about nurturing their mental health and confidence. & Happiness \\
&(7) We need to create a supportive environment where kids can grow freely instead of being bound by invisible pressure. & Neutral \\
\bottomrule
\end{tabular}
\caption{Representative dialogue examples from different domains in the self-constructed dataset (Part I).}
\label{dataset_examples_part1}
\end{table*}

\begin{table*}[htbp]
\centering
\begin{tabular}{l|p{12cm}c}
\toprule
Domain & Example Utterance & Emotion\\
\midrule
Family 
&(1) Lately, it feels like there's a tense atmosphere at home. No one's really talking, and honestly, it makes me feel scared. & Fear \\
&(2) Same here. Every time I get home, I just see my parents looking down at their phones, and it really hurts. It feels like we've lost our bridge to communicate. & Sadness \\
&(3) Exactly. If we don't talk things out soon, the issues will just get worse. We should find a good time to sit down and have a real conversation, but I'm always worried it'll turn into an argument. & Fear \\
&(4) Me too. Every time family issues come up, the mood just gets bad, and it feels like everyone is avoiding it. & Sadness \\
&(5) But if we don't talk about it, the frustration just builds up inside. It's honestly such a sad feeling. Are we just going to stay silent forever? & Sadness \\
&(6) That's why I think we need to face this, even if it's hard. We have to find the right moment and try to gently encourage everyone to open up. If we keep going like this, it'll really lead to despair. & Anger \\
\midrule
Social 
&(1) It feels really lonely; everyone else seems so relaxed, but I just can't seem to fit in. & Sadness \\
&(2) Whenever I think about saying hello, my mind fills with all sorts of negative thoughts. & Fear \\
&(3) But then I think, if I could just muster the courage to chat, I might find unexpected joy. & Happiness \\
&(4) Yeah, maybe we can practice together, support each other, and slowly overcome this anxiety. & Happiness \\
&(5) Exactly! That way, we won't feel so scared anymore. & Happiness \\
\midrule
Entertainment 
&(1) I watched that new horror movie last night, and it was terrifying! I couldn't help but cover my eyes during the tense scenes. & Fear \\
&(2) Same here! Some scenes made me think about things that could actually happen in real life, which was really creepy. & Fear \\
&(3) Exactly! Especially the parts about stalkers—I started worrying about whether someone might be following me home at night. & Fear \\
&(4) That's the part I hate the most! The unease from the movie really lingers into reality. I even left the lights on while I was trying to sleep last night. & Fear \\
&(5) Right? I started doubting the people around me, thinking they might be hiding something. & Fear \\
&(6) We are really sensitive, but there's something thrilling about this fear. & Neutral \\
&(7) Sometimes, I actually find the experience fun. Even though I'm scared, I still want to watch it again. & Happiness \\
&(8) Maybe we should find some lighter movies to relax a bit, but the allure of horror films is just irresistible! & Happiness \\
\bottomrule
\end{tabular}
\caption{Representative dialogue examples from different domains in the self-constructed dataset (Part II).}
\label{dataset_examples_part2}
\end{table*}

\section{Appendix B}
\label{appendix:implementation}
The model is fine-tuned using LoRA with a rank of 32. The learning rate is set to 3e-4, and the optimizer is paired with a linear learning rate scheduler. Training runs for a total of 4 epochs, with curriculum learning applied during the first two epochs, followed by two epochs of overall training. The curriculum learning strategy divides the training data into two difficulty levels and updates the curriculum level at the end of each epoch. The context window size is set to 5 to enable the model to effectively integrate preceding context. Training uses a batch size of 1 with 4 steps of gradient accumulation. The maximum number of training steps is controlled by a parameter. The maximum gradient norm is set to 0.3, and a warm-up ratio of 0.03 is applied. Evaluation, model saving, and logging are performed every 50 steps, using the weighted F1 score as the metric to select the best model.

\end{document}